\pdfoutput=1
\documentclass{article}

\PassOptionsToPackage{numbers}{natbib}

 \usepackage[preprint]{neurips_2019}

\usepackage[utf8]{inputenc} 
\usepackage[T1]{fontenc}    
\usepackage{hyperref}       
\usepackage{url}            
\usepackage{booktabs}       
\usepackage{amsfonts}       
\usepackage{nicefrac}       
\usepackage{microtype}      
\usepackage{hyperref}

\usepackage{color,scalefnt,soul,epsfig}
\usepackage{caption}
\usepackage{subcaption}
\usepackage{amsmath}

\DeclareMathOperator{\pool}{pool}
\usepackage{graphicx} 
\usepackage{ragged2e}

\usepackage{amssymb}
\usepackage{tikz}
\usetikzlibrary{shapes,arrows,positioning,matrix}
\usepackage{pgfplotstable}
\usepackage{pgfplots}

\title{OrderNet: Ordering by Example}

\author{%
    Robert A.~Porter\\
    GradeCam\\
    Livermore, CA\\
  \texttt{physicsrob@gmail.com} \\
}

\begin{document}

\maketitle

\begin{abstract}
In this paper we introduce a new neural architecture for sorting unordered sequences where the correct sequence order is not easily defined but must rather be inferred from training data.  We refer to this architecture as OrderNet and describe how it was constructed to be naturally permutation equivariant while still allowing for rich interactions of elements of the input set.  We evaluate the capabilities of our architecture by training it to approximate solutions for the Traveling Salesman Problem and find that it outperforms previously studied supervised techniques in its ability to generalize to longer sequences than it was trained with.  We further demonstrate the capability by reconstructing the order of sentences with scrambled word order.

\end{abstract}

\section{Introduction}
In this paper we examine the problem of ordering sets with neural networks.  We use ordering to mean any function that takes as input an unordered set and outputs the elements of the input in a specific order.

Using this definition, we can break ordering into three non-comprehensive categories:
\begin{enumerate}
    \item \textbf{\textit{Easy:}}  If an ordering is fully specified by a well-defined comparison operator $\leq$ between input elements, the problem is easy.  This is just the traditional task of sorting a list, and there are a plethora of sorting algorithms that can easily determine the correct order. \\
        \textbf{Example:} Given a list of numbers, sort them from smallest to largest. \\
    \item \textbf{\textit{Harder:}}  If an ordering is fully specified by a comparison of input elements but we have incomplete knowledge, the problem is harder. \\
        \textbf{Example:} Given a list of facial images, sort them from the youngest person to the oldest person.\\
    \item \textbf{\textit{Hardest:}}  If an ordering is only determined by considering every element of the input, the problem is hard.  These are combinatorial optimization problems, but the goal we are optimizing for might be probabilistic in nature, e.g. What is the most probable order for the input? \\
        \textbf{Example:} Take an example sentence and scramble the order of the words, e.g. [“example”, “is”, “This”, “an”, “.”].  Can we reconstruct the original sentence, “This is an example.”?
\end{enumerate}

The easy category of ordering problems never benefits from the application of machine learning.  The harder category likely can benefit from machine learning, but it is probably more fruitful to use machine learning to infer an intermediate value that can be sorted using an ordinary sort algorithm.

We are interested in solving the hardest case:  Given training pairs $(X,Y)$ where each $X$ is an unordered input and each $Y$ is the observed ordering of the input, we wish to build a model $P(Y | X)$ that allows us to autoregessively predict the most probable ordering.

Sequence-to-sequence models \cite{sutskever2014sequence} (seq2seq) have become ubiquitous in many areas of deep learning including language translation, and one might initially think they may be a good fit for modeling $P(Y|X)$ in ordering problems.  Indeed, one could think of ordering as the task of translating an unordered set into an ordered sequence, but there are two aspects which make seq2seq inappropriate:
\begin{enumerate}
\item The outputs of the model are indices which reference elements of the input.  If a fixed output dictionary is used, as is usually the case in seq2seq models, the model would not be able to generalize to longer sequence lengths.
\item The input does not have a known ordering, and thus there is a large equivalence class.  Since seq2seq models are not naturally permutation equivariant, the model would need to learn the ordering for each of $n!$ permutations.
\end{enumerate}

The first problem has been addressed previously using Pointer Networks \cite{vinyals2015pointer}, which combine an LSTM \cite{hochreiter1997long} seq2seq model with an attention mechanism, inspired by \cite{bahdanau2014neural}, such that the output sequence can learn to point at elements of the input sequence.  We draw heavy inspiration from that work; however, it leaves the problem of the large equivalence class of the input set unaddressed.  In a follow up work Vinyals et al \cite{vinyals2015order} address the second problem of permutation equivariance by encoding the input set into a memory block which is addressed with a content-based attention mechanism.

In this paper we present a new neural architecture for ordering unordered sets where the order is inferred from data.  We refer to this model as OrderNet, and we demonstrate its capabilities by applying it to the Traveling Salesman Problem (TSP) in Section~\ref{sec:tsp}.  We show that it is more capable of generalizing to longer sequences than existing techniques and uses fewer than half the parameters.  In Section~\ref{sec:wordorder} we apply OrderNet to reconstruct the word order of
randomly shuffled Wikipedia text.

\paragraph{Main Contributions} The main contributions of this paper are as follows:
\begin{itemize}
\item An encoder for sets which is more expressive than some previously studied methods.
\item A convolutional decoder for set2seq problems that maintains permutation equivariance.
\item A generic end-to-end model for learning to reconstruct the order of a sequence.
\end{itemize}


\section{Model}
\paragraph{Notation} Input elements are indexed with the symbol $i$ with values we represent as A,B,C,...  We use alpha characters to emphasize that the input is an unordered set.  Output sequence elements are indexed with the symbol $t$ with values $0,1,2,...$.  We use $[a \;\; b]$ to denote vector concatenation of $a$ with $b$, and we often drop indices representing the feature dimension of a vector.

\paragraph{Problem Formulation}
Given training data which consists of example pairs $(X, Y)$, where each $X$ is an unordered set of vectors $x_i \in \mathbb{R}^d$ , and each $Y$ is a target sequences of indices $y_t \in \mathbb{N}$ representing the order of $X$, we desire to model the conditional probability $P(Y | X)$.  In the case of the Traveling Salesman Problem, which we visit in Section~\ref{sec:tsp}, each $X$ would be a set of cities to be visited (2-dimensional random points in the interval $[0,1]$), and each corresponding $Y$ would be the sequence of indices which result in the shortest possible path that visits each city exactly once.

Much as in the case of seq2seq modeling, we can decompose this probability without loss of generality using the chain rule:

\begin{equation}\label{eqn:condprob}
    P (Y | X ) = \prod_{t} P ( y_t | y_1, \ldots, y_{t-1}, X )
\end{equation}

Formulating the probability using the chain rule allows us to construct our model autoregressively.  We must only model the probability of a single timestep conditioned on the input set and the previously predicted output.

We formulate our model as an encoder-decoder: First, the input is passed through an encoder which allows the model to capture context for each element of the input.  Second, the encoded representation of the input set is passed to a decoder which models the conditional probability.

Our method differs from typical seq2seq approaches in two important ways:
\begin{enumerate}
\item The encoder output size is not fixed.  Each element of the input set is encoded to capture context.  The encoder output is fixed-size for each element.  The encoding is constructed to be permutation equivariant.  
\item The decoder output is a softmax probability distribution over input elements, similar to \cite{vinyals2015pointer}.
\end{enumerate}

\paragraph{Permutation Equivariance} Ordering functions must be equivariant to permutations.  In the case of the TSP, if the first two cities are swapped in the input, then the values $1$ and $2$ must be exchanged in the output.  We could learn this equivariance from data; however, there are $n!$ permutations for each input set, so doing so would be grossly inefficient with model parameters, and generalization would most likely suffer.  Instead we choose to construct each component of our model with permutation equivariance in mind.

\subsection{Input Set Encoder}
The first component of OrderNet is an encoder which captures context for each input set element.  The goal of the encoder is to map each element $x \in X$ into a representation which will make decoding the set into the correct order easier.

Without loss of generality, any such representation can be decomposed as a function $\mathrm{Encoding}(x_i)=f(x_i,X)$, and the encoding will be permutation equivariant so long as the function's dependence on the set $X$ is invariant to permutations.

For our encoder we limit our consideration to functions of the form $f(x_i,X) = \pool_j \phi(x_i, x_j)$, where $\pool$ is either max pooling or average pooling, and we use a two-layer MLP for $\phi$.  This construction has several benefits:
\begin{itemize}
    \item Pooling over all $j$ ensures permutation equivariance.
    \item Using a two-layer MLP allows $\phi$ to approximate any arbitrary function, limited only by the width of the layers \cite{cybenko1989approximation}.
    \item By representing the encoding as a function of all pairs of input elements we gain expressive capacity.
\end{itemize}

We refer to this sequence of operations as the OrderNet encoder block, and we construct the encoder by applying several layers of encoder blocks in sequence. 
The input of each block is concatenated
with its output so that the final representation contains all information available from previous encoding layers,
including the original input. We conjecture that multiple layers allow for a much richer representation.
Roughly speaking, the first layer allows for pairwise interactions, the second layer allows for pairwise
interactions of pairs, so on and so forth. Each additional layer allows for more interactions between
the various subsets of the input.

Although the use of pooling operators for encoding sets has been studied extensively in other papers, it is typically applied to functions of input elements \cite{zaheer2017deep} \cite{qi2017pointnet}, not functions of pairs of elements.  Our encoder is more expressive since we allow each input element to interact with every other element of the input set before pooling.  It is also worth noting that the way we have constructed the encoder bears some resemblance to self-attention mechanisms
studied extensively in some language translation models \cite{vaswani2017attention}.


\paragraph{Encoder Implementation}
The input to OrderNet is represented as a matrix $X_{id}$ where $i$ indexes the element of the set, and $d$ indexes the feature dimension of the vector elements of the set.
In order to perform our encoding operation where each element interacts with every other element, we must first expand and concatenate the input $X_{id}$ with a transposed version of itself to form a 3D tensor $Z \in \mathbb{R}^{|X| \times |X| \times 2d}$, where

\begin{equation}
    Z_{ij} = [x_{i} \;\; x_{j} ]
\end{equation}

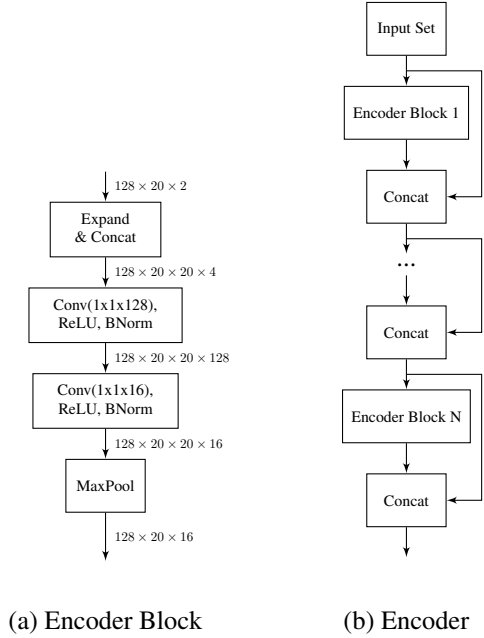
\begin{figure}
    \centering

    \tikzstyle{block} = [draw, fill=white, rectangle, minimum height=2em, minimum width=3em]
    \tikzstyle{sum} = [draw, fill=white, circle, node distance=1cm]
    \tikzstyle{input} = [coordinate]
    \tikzstyle{output} = [coordinate]
    \tikzstyle{pinstyle} = [pin edge={to-,thin,black}]

    \begin{tikzpicture}[auto, node distance=0.4cm,>=latex']
        \node [input, name=input] at (0,0) {};
        \node [block, below=of input, node distance=1cm] (expand) {\tiny \begin{tabular}{c} Expand \\ \& Concat\end{tabular} };
        \node (conv1) [block, below=of expand] {\tiny \begin{tabular}{c} Conv(1x1x128), \\ ReLU, BNorm  \end{tabular}};
        \node (conv2) [block, below=of conv1] {\tiny \begin{tabular}{c} Conv(1x1x16), \\ ReLU, BNorm  \end{tabular}};
        \node (maxpool) [block, below=of conv2] {\tiny MaxPool};
        \node (output) [output, below of=maxpool, node distance=1cm] {};
        
        \draw [draw,->] (input) -- (expand) node [midway, right] (dim1) {\scalebox{0.5} {$128\times20\times2$}};
        \draw [draw,->] (expand) -- (conv1) node [midway, right] (dim2) {\scalebox{0.5} {$128\times20\times20\times4$}};
        \draw [draw,->] (conv1) -- (conv2) node [midway, right] (dim3) {\scalebox{0.5} {$128\times20\times20\times128$}};
        \draw [draw,->] (conv2) -- (maxpool) node [midway, right] (dim3) {\scalebox{0.5} {$128\times20\times20\times16$}};
        \draw [draw,->] (maxpool) -- (output) node [midway, right] (dim4) {\scalebox{0.5} {$128\times20\times16$}};

        \node [block] (inp) at (4,1.9) {\tiny Input Set};
        \node [block, below=of inp] (attn1) {\tiny Encoder Block 1};
        \node [block, below=of attn1] (concat1) {\tiny Concat};
        \node [below=of concat1] (attn2) {...};
        \node [block, below=of attn2, node distance=3] (concat2) {\tiny Concat};
        \node [block, below=of concat2] (attn3) {\tiny Encoder Block N};
        \node [block, below=of attn3] (concat3) {\tiny Concat};
        \node [output, name=out, below=of concat3, node distance=1cm] {};

        \draw [draw,->] (inp) -- (attn1) coordinate [midway] (edge1) {};
        \draw [draw,->] (attn1) -- (concat1);
        \draw [draw,->] (concat1) -- (attn2) coordinate [midway] (edge2) {};
        \draw [draw,->] (attn2) -- (concat2);
        \draw [draw,->] (concat2) -- (attn3) coordinate [midway] (edge3) {};
        \draw [draw,->] (attn3) -- (concat3);
        \draw [draw,->] (concat3) -- (out);
        
        \draw [draw,->] (edge1) -- ([xshift=1cm]edge1.center) -- ([xshift=1cm]concat1.center) -- (concat1);
        \draw [draw,->] (edge2) -- ([xshift=1cm]edge2.center) -- ([xshift=1cm]concat2.center) -- (concat2);
        \draw [draw,->] (edge3) -- ([xshift=1cm]edge3.center) -- ([xshift=1cm]concat3.center) -- (concat3);

        \node [] at (0,-6) {(a) Encoder Block};
        \node [] at (4,-6) {(b) Encoder};

    \end{tikzpicture}
    \caption{OrderNet Encoder.  (a) Depiction of the first Encoder Block.  Size of tensor for the TSP example are indicated to the right of the edges (assuming 20 cities and batch size of 128). (b) Encoder is built by stacking N Encoder Blocks on top of each other, with outputs concatenated after each block.}
    \label{fig:encoder}
\end{figure}

To implement the two-layer MLP we apply a sequence of $1\times1$ convolution, ReLU \cite{nair2010rectified}, and batch normalization \cite{ioffe2015batch}, repeated once. Following the two-layer MLP, the tensor is pooled over the $j$ dimension yielding an encoding of $X_{id}$ which captures the context of the set.  The encoder block is depicted in Figure~\ref{fig:encoder} (a).  It should be noted that the $1\times1$ filter size of the convolutions is chosen specifically to preserve permutation equivariance as using a larger filter size would allow the network to learn dependencies between adjacent elements of the input.  We set the diagonal elements to zero before pooling with the motivation of preventing the self-interaction of elements from dominating the encoding.

\subsection{Fully Convolutional Decoder}
Taking inspiration from previously developed fully convolutional encoder-decoder models \cite{elbayad2018pervasive}, the OrderNet decoder is fully convolutional, avoiding recurrent networks and allowing for training to happen for each timestep in parallel.  The inputs to the decoder are elements of the input set as represented by the output of the encoder module.  The decoder is autoregressive: we make predictions for each timestep $t$ that depend on the prediction from timestep $t-1$
\cite{graves2013generating}.

We define $x'_i$ to be the encoder representation of element $i$ from the input set with feature depth $d'$ and refer to these as the source elements.
We define $y'_t$ to be the encoder representation of the previously decoded output for timestep $t-1$ and refer to these as target elements.
By expanding and concatenating the source and target elements, we can define a 3D source-target tensor $S_{it}$, where $i$ indexes the elements of the input set and $t$ indexes the output timestep.  $S \in \mathbb{R}^{|X| \times |X| \times 2d'}$, where

\begin{align*}
x'_i &= \text{Encoder}(x_i, X), \\
y'_t &= \text{Encoder}(x_{y_{t-1}}, X) \text{ for } t>0, \\
y'_0 &= \text{<start token>} \text{, and}\\
S_{it} &= [x'_i \;\; y'_t].\\
\end{align*}

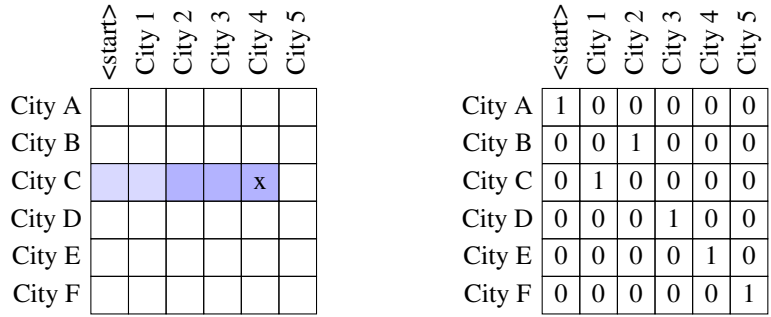
\begin{figure}
    \centering
    \begin{tikzpicture}[cell/.style={rectangle,draw=black}]
        \matrix (convs) at (0,0) [
            matrix of nodes,
            row sep =-\pgflinewidth,
            column sep = -\pgflinewidth,
            execute at empty cell={\node{\strut};},
            nodes={cell,anchor=center,minimum width=0.5cm,text height=0.5em,minimum height=0.5cm,text depth=0em},
            column 1/.style = {nodes={cell, draw=white, minimum width=1cm}},
            row 1/.style={nodes={cell,draw=white,rotate=90,text height=0.5em,text depth=0em}},
        ]
        {
            & <start> & City 1 & City 2 & City 3 & City 4 & City 5\\
            City A   &  &  &  &  & & \\
            City B   &  &  &  &  & & \\
            City C   & |[fill=blue!15]| & |[fill=blue!15]| & |[fill=blue!30]| & |[fill=blue!30]| & |[fill=blue!30]| x & \\
            City D   &  &  &  &  & & \\
            City E   &  &  &  &  & & \\
            City F   &  &  &  &  & & \\
        };
        \node [below=of convs, text width=6cm] {\small(a) Receptive field of convolutions};

        \matrix (target) at (6,0) [
            matrix of nodes,
            row sep =-\pgflinewidth,
            column sep = -\pgflinewidth,
            execute at begin cell=\strut,
            execute at empty cell={\node{\strut};},
            nodes={cell,anchor=center,minimum width=0.5cm,text height=0.5em,text depth=0em, minimum height=0.5cm},
            column 1/.style = {nodes={cell, draw=white, minimum width=1cm}},
            row 1/.style={nodes={cell,draw=white,rotate=90,text height=0.5em,text depth=0em}},
        ]
        {
            & <start> & City 1 & City 2 & City 3 & City 4 & City 5\\
            City A   & 1 & 0 & 0 & 0 & 0 & 0\\
            City B   & 0 & 0 & 1 & 0 & 0 & 0\\
            City C   & 0 & 1 & 0 & 0 & 0 & 0\\
            City D   & 0 & 0 & 0 & 1 & 0 & 0\\
            City E   & 0 & 0 & 0 & 0 & 1 & 0\\
            City F   & 0 & 0 & 0 & 0 & 0 & 1\\
        };
        \node [below=of target] {\small(b) Target values};
    \end{tikzpicture}

    \caption{\textbf{(a)} \textit{Receptive field of convolutions -} Illustration of receptive field for City C, timestep 4 (marked x).  Blue represents receptive field after first $1\times3$ convolution; light blue represents receptive field after second $1\times3$ convolution.  Note the causal padding prevents the receptive field from seeing into the future.  \textbf{(b)} \textit{Target values -} Target value for output of decoder.  A cross-entropy loss is used for each timestep.};
    \label{fig:convs}
\end{figure}

\begin{figure}
    \centering
    \tikzstyle{block} = [draw, fill=white, rectangle, minimum height=2em, minimum width=3em]
    \tikzstyle{input} = [coordinate]
    \tikzstyle{output} = [coordinate]
    \tikzstyle{pinstyle} = [pin edge={to-,thin,black}]
    
    \begin{tikzpicture}[auto, node distance=0.5cm,>=latex']
        \node [input, name=input] at (0,0) {};
        \node [block, below=of input, node distance=1cm] (conv) {\tiny $1\times3$ Conv};
        \node [block, below=of conv] (relu) {\tiny ReLU};
        \node [block, below=of relu] (bnorm) {\tiny BNorm};
        \node [block, below=of bnorm] (concat) {\tiny Concat};
        \node [output, name=output, below=of concat, node distance=1cm] {};
        
        \node [block, left=of conv] (maxpool) {\tiny MaxPool};
        
        \draw [draw,->] (input) -- (conv) coordinate [midway] (inpEdge) {};
        \draw [draw,->] (conv) -- (relu);
        \draw [draw,->] (relu) -- (bnorm);
        \draw [draw,->] (bnorm) -- (concat);
        \draw [draw,->] (concat) -- (output);
        \draw [draw,->] (inpEdge) -| (maxpool);
        \draw [draw,->] (maxpool) |- (concat);

        \draw [draw,->] (inpEdge) -- ([xshift=1cm]inpEdge.center) -- ([xshift=1cm]concat.center) -- (concat);

        \node [block] (stinp) at (4,-0.5) {\tiny \begin{tabular}{c} Source-Target \\ Tensor\end{tabular}};
        \node [block, below=of stinp, node distance=1cm] (block1) {\tiny Decoder Block 1};
        \node [block, below=of block1, node distance=1cm] (block2) {\tiny Decoder Block 2};
        \node [block, below=of block2, node distance=1cm] (fconv) {\tiny $1\times1$ Conv};
        \node [block, below=of fconv, node distance=1cm] (softmax) {\tiny Softmax};
        \node [output, name=decout, below=of softmax, node distance=1cm] {};
        \draw [draw,->] (stinp) -- (block1);
        \draw [draw,->] (block1) -- (block2);
        \draw [draw,->] (block2) -- (fconv);
        \draw [draw,->] (fconv) -- (softmax);
        \draw [draw,->] (softmax) -- (decout);
        
        \node [] at (0,-7) {(a) Decoder Block};
        \node [] at (4,-7) {(b) Decoder};
       %
%

    \end{tikzpicture}
    \caption{OrderNet Decoder: (a) an individual decoder block, (b) multiple decoder blocks are stacked on top of each other to form the decoder.}
    \label{fig:decoder}
\end{figure}
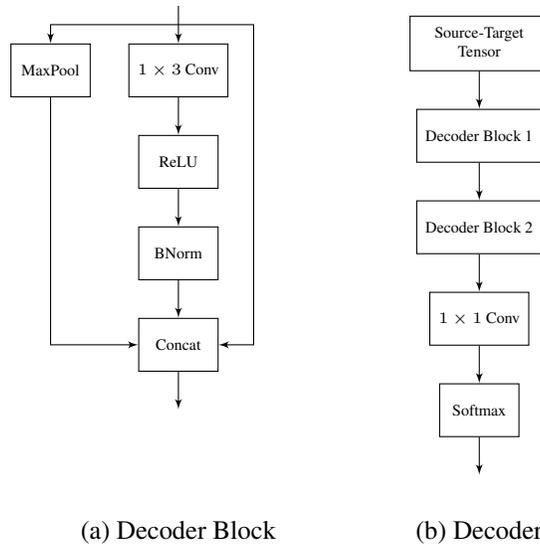

An illustration of the source-target tensor is provided in Figure~\ref{fig:srctarget}.  We build our decoder from layers of convolutions, non-linearities (ReLU), batch normalizations, and max pooling.  We construct the decoder with the following in mind:
\begin{itemize}
    \item Causal padding is necessary to prevent the layers from seeing into the future (see Figure~\ref{fig:convs}).
    \item To preserve permutation equivariance, the filter size must be $1$ in the source-dimension to prevent input elements from interacting.
    \item Max pooling in the source-dimension allows for the decoder to see other possible decodings without breaking permutation equivariance (as long as the pooling spans the entirety of the source-dimension).
    \item Dense connectivity has been demonstrated to improve the performance of deep networks \cite{huang2017densely}, so with that in mind the input to each decoder block is concatenated with its output.
\end{itemize}
The configuration of layers used for the decoder is shown in Figure~\ref{fig:decoder}.

\begin{figure}
    \centering
    \begin{tikzpicture}[cell/.style={rectangle,draw=black}]

        \matrix (values) [
            matrix of nodes,
            row sep =-\pgflinewidth,
            column sep = -\pgflinewidth,
            execute at begin cell=\strut,
            execute at empty cell={\node{\strut};},
            nodes={cell,anchor=center,minimum width=1.5cm,minimum height=1cm},
            column 1/.style = {nodes={cell, draw=white, minimum width=1cm}},
            row 1/.style={nodes={cell,draw=white,rotate=90,text height=0.5em,text depth=0em}},
        ]
        {
            & <start> & City 1 & City 2 & City 3 & City 4 & City 5\\
            City A   & $[x'_A\;\;y'_0]$ & 0  & 0 & 0 & 0 & 0 \\
            City B   & $[x'_B\;\;y'_0]$ & $[x'_B\;\;y'_1]$  & $[x'_B\;\;y'_2]$ & 0 & 0 & 0\\
            City C   & $[x'_C\;\;y'_0]$ & $[x'_C\;\;y'_1]$  & 0 & 0 & 0 & 0\\
            City D   & $[x'_D\;\;y'_0]$ & $[x'_D\;\;y'_1]$  & $[x'_D\;\;y'_2]$ & $[x'_D\;\;y'_3]$  & 0 & 0 \\
            City E   & $[x'_E\;\;y'_0]$ & $[x'_E\;\;y'_1]$  & $[x'_E\;\;y'_2]$ & $[x'_E\;\;y'_3]$  & $[x'_E\;\;y'_4]$ & 0\\
            City F   & $[x'_F\;\;y'_0]$ & $[x'_F\;\;y'_1]$  & $[x'_F\;\;y'_2]$ & $[x'_F\;\;y'_3]$  & $[x'_F\;\;y'_4]$ & $[x'_F\;\;y'_5]$ \\
        };
        \node [above=of values-1-4, anchor=center] {Target Sequence};
        \node [left=of values-4-1, anchor=center, rotate=90] {Source Set};
    \end{tikzpicture}

    \caption{Illustration of source-target tensor.  This tensor is the input to the decoder.  The example illustration is for the 6-city TSP case where the shortest path is $A \rightarrow C\rightarrow B \rightarrow D \rightarrow E \rightarrow F \rightarrow A$.  Each cell is a vector concatenation of the self-attention representation for the source and target.  Once a city has been visited, we set the value to zero for all subsequent timesteps.}

    \label{fig:srctarget}
\end{figure}

\subsection{Hyperparameters}
OrderNet has several hyperparameters, including the depths of the individual layers in the encoder and decoder blocks, the number of blocks, and the pooling operator used for the encoder.  Although an extensive hyperparameter search was not performed, we did experiment with several options.  The hyperparameters used for the experiments in Section~\ref{sec:exp} are reported in Table~\ref{table:hyper}.

\begin{table}
  \caption{Hyperparameters used in experiments}
\label{table:hyper}
  \centering
  \begin{tabular}{lll}
    \toprule
        Hyperparameter & TSP Experiment & Word Order Experiment\\
    \midrule
        Encoder Blocks & 4 & 8 \\
        Encoder Layer-1 Depth & 128 & 256 \\
        Encoder Layer-2 Depth & 16 & 32 \\
        Encoder Pooling & MaxPool & AvgPool \\
        Decoder Blocks & 4 & 8 \\
        Decoder Block Depth & 16 & 32\\
    \midrule
        Parameters & 73K & 1.4M \\
    \bottomrule
  \end{tabular}
\end{table}

\section{Experiments}
\label{sec:exp}
\subsection{Traveling Salesman Problem}
\label{sec:tsp}
The Traveling Salesman Problem is described as follows: Given a set of cities and a starting city, find the shortest route for a salesman to visit each city exactly once and then return to the starting city.  In our case, we focus on the TSP in 2D Euclidian space, where each city is a random point in the interval $[0,1]$.
This problem has been studied extensively.  A brute-force solution requires calculating the path length for $n!$ permutations, which becomes impractical very quickly.  
In 1962 Held and Karp \cite{held1962dynamic}, and independently Bellman \cite{bellman1961dynamic}, developed an algorithm using dynamic programming to find the shortest path with time complexity $O(2^n n^2)$.  In 1976 Christofides \cite{christofides1976worst} developed a heuristic-based approach which can find a solution guaranteed to be at most $1.5 \times$ longer than the optimal path in $O(n^3)$ time.  
More recently the idea of using machine learning to solve combinatorial optimization problems with reinforcement learning has been explored \cite{kool2018attention} \cite{bello2016neural} \cite{khalil2017learning}; however, we note that the goal of OrderNet is not to find good solutions to well-defined combinatorial optimization problems but rather to infer orderings from example data.  We use the TSP to study the ability of our model to infer the correct order of a set of vectors.

In our framing, the cities in the TSP problem are just $2D$ vectors with an unknown order.  We use supervised learning to train our model to imitate the ordering observed in examples of exact solutions that minimize the total path length.  For each number of cities in the range 5 to 20, we generated 100,000 training pairs using random city locations and the Held-Karp method of finding the order which yields the shortest path.  We train the model with Adam \cite{kingma2014adam} by minimizing the cross-entropy
loss between the predicted and observed order.  Following training for 10 epochs, we find that our model has converged and produces surprisingly good solutions.  A beam search with a beam size of $5$ is used to predict the most likely path.

\begin{figure}
    \centering
    \begin{tikzpicture}
    \begin{axis}[
        xmin=-15,xmax=105,
        ymin=1,ymax=10,
        xlabel=Number of Cities,
        ylabel=Tour Length,
        legend style={at={(0,1)},anchor=north west}
      ]
    \addplot [red, mark=square*, mark size=1pt] table [y=L, x=N]{christofides.dat};
    \addlegendentry{Christofides}
    \addplot [blue, mark=square *, mark size=3pt, line width=2pt] table [y=L, x=N]{exact_tsp.dat};
    \addlegendentry{Exact}
    \addplot [orange, mark=square*, mark size=1pt] table [y=L, x=N]{ptrnet_tsp.dat};
    \addlegendentry{Pointer Networks} 
    \addplot [green!60!black, mark=square*, mark size=1pt] table [y=L, x=N]{our_tsp.dat};
    \addlegendentry{OrderNet}
    \end{axis}
    \end{tikzpicture}
    \caption{Traveling Salesman average tour length (lower is better): Christofides (approximate heuristic solver with time complexity $O(n^3)$), Exact solutions obtained with the Held-Karp method (time complexity $O(2^n n^2)$), Pointer Networks (time complexity $O(n^2)$) tour lengths as reported in \cite{vinyals2015pointer}, and our model OrderNet (time complexity $O(n^2)$).  Both pointer networks and our model were trained using exact solutions for tour lengths between $5-20$.  All methods provide near-exact results in the interval $N \in [5,20]$.}
    \label{fig:tsp}
\end{figure}
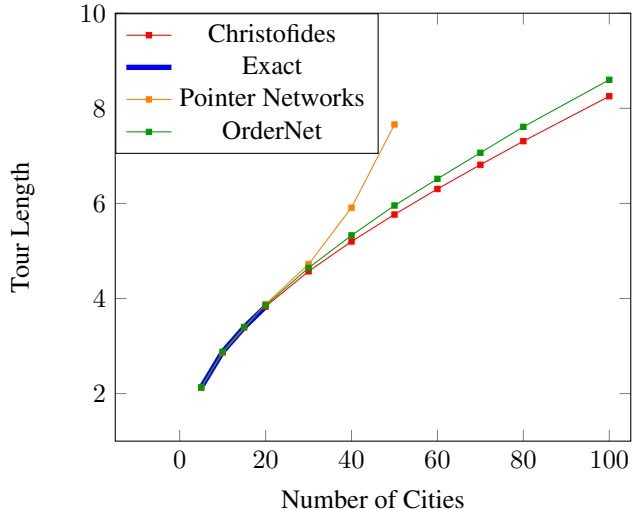

Figure~\ref{fig:tsp} shows the average length of the proposed tours our model generates as a function of the number of cities.  Not only is the model capable of reproducing high quality tours for a similar number of cities as our training data, but also our model is able to generalize and produce good solutions for tours at least 5 times longer than those on which it was trained.  In the
cases we studied, tours of up to 100 cities, the tours generated by our model are only slightly worse than the tours produced using the Christofides algorithm, despite our model having time complexity $O(n^2)$ and Christofides having time complexity $O(n^3)$.

\subsection{Word Order}
\label{sec:wordorder}
In this section we analyze an example problem of learning an order: Given the first 5 words of a line of Wikipedia text, lower cased and in randomized order, can we infer the correct order?  
To examine this problem we use the WikiText-103 dataset \cite{merity2016pointer}, processed to only consider the first five words of text for each line, and disregarded blank lines and section headers.  The resulting dataset had $597,867$ training examples of 5-word sequences.  We encode each word using pretrained GloVe \cite{pennington2014glove} embeddings (50-dimensional), and feed them to OrderNet.  We find that OrderNet predicts the correct word order $69.5\%$ of the time.
A selection of 10 sequences is presented in Table~\ref{table:wikiexamples}.  The majority of mistakes encountered represent ambiguity in the data as formulated.  Without additional context it is not necessarily possible to accurately reconstruct the order of these words.

\begin{table}
  \caption{WikiText-103 samples with actual and predicted word order.  Incorrect predictions are highlighted in light-red.}
  \label{table:wikiexamples}
  \centering
  \begin{tabular}{lll}
    \toprule
    Actual Word Order & Predicted Word Order\\
    \midrule
    <unk> pike as miranda frost & \colorbox{red!10}{miranda frost as <unk> pike} \\
    while deangelo gets annoyed with & while deangelo gets annoyed with \\
    the tna x division championship & the tna x division championship \\
    federal bureau of investigation agents & federal bureau of investigation agents  \\
    as the japanese regrouped west & as the japanese regrouped west \\
    meridian has nine historic districts & \colorbox{red!10}{historic meridian has nine districts} \\
    when drummer lombardo left slayer & when drummer lombardo left slayer \\
    the spores are 7 – & the spores are 7 – \\
    highland park houses a jimmie & \colorbox{red!10}{jimmie houses a highland park} \\
    route 29 follows main street & \colorbox{red!10}{main street follows route 29} \\
    \bottomrule
  \end{tabular}
\end{table}

\section{Conclusion}
The goal of this paper was to build a neural network capable of reconstructing the order of sequences.  Such a network must be permutation equivariant, which places strong constraints on the way it can be constructed.  Regardless of these constraints, we were able to build an expressive neural network capable of modeling complicated interactions of input elements.  The results for the Traveling Salesman Problem show that OrderNet is capable of extrapolating to
sequence lengths far greater than those on which it was trained.  The results for the word order problem demonstrate that OrderNet is applicable to a broad range of problems.  In future work we are interested in additional applications of OrderNet, including its components.  In particular the OrderNet encoder may have applications to other problems where the function being modeled must be equivariant (or invariant) to the input order.

\bibliographystyle{unsrt}
\bibliography{main}

\end{document}